\documentclass[11pt]{article}
\usepackage{eamt18}
\usepackage{times}
\usepackage{url}
\usepackage{latexsym}
\usepackage{ctable}
\usepackage[title]{appendix}
\usepackage{bm}
\usepackage{soul}
\usepackage[small,bf]{caption} 
\title{Deep Neural Machine Translation with Weakly-Recurrent Units}

\author{
  Mattia A. Di Gangi \\
  FBK, Trento, Italy \\
  University of Trento, Italy \\
  {\tt digangi@fbk.eu} \\\And
  Marcello Federico \\
  MMT Srl, Trento, Italy \\
  FBK,  Trento, Italy \\
  {\tt federico@fbk.eu} \\}

\date{}

\begin{document}
\maketitle
\begin{abstract}
Recurrent neural networks (RNNs) have represented for years the state of the art in neural machine translation.
Recently, new architectures have been proposed, which can  leverage parallel computation on GPUs better than classical RNNs.
Faster training and inference combined with different sequence-to-sequence modeling also lead to performance improvements.
While the new models completely depart from the original recurrent architecture, we decided to investigate how to make RNNs more efficient.
In this work, we propose a new recurrent NMT architecture, called Simple Recurrent NMT, built on a class of fast and weakly-recurrent units that use layer normalization and multiple attentions. 
Our experiments on the WMT14 English-to-German and WMT16 English-Romanian benchmarks show that our model represents a valid alternative to LSTMs, as it can achieve better results at a significantly lower computational cost.
\end{abstract}

\section{Introduction}

Neural machine translation (NMT)~\cite{sutskever2014sequence,bahdanau2014neural} is a sequence-to-sequence problem that requires generating a sentence in a target language from a corresponding sentence in a source language. 
Similarly to other language processing task, NMT has mostly employed recurrent neural networks (RNNs)~\cite{sennrich2016edinburgh,sennrich2017university,luong2015stanford}, in both their LSTM~\cite{hochreiter1997long} and GRU~\cite{cho2014properties} variants, which can model long-range dependencies. Besides their simplicity, 
the choice of RNNs is also due to their expressive power, which has been proven equivalent to Turing Machines~\cite{siegelmann1995computational}.
RNNs have represented so far the state of the art of machine translation, and have constantly been enhanced to improve their performance.
Nonetheless, their explicit time dependencies make training of deep RNNs computationally very expensive \cite{wu2016google,barone2017deep}.

Recent works have proposed new NMT architectures, not based on RNNs, that obtained significant improvements both in training speed and translation quality: the so-called convolutional sequence-to-sequence ~\cite{gehring2017convolutional} and the self-attentive or transformer  \cite{vaswani2017attention} models. 
Speed improvements by these models mainly come from the possibility of parallelizing computations over word sequences, as both models do not have time dependencies. 
On the other hand, performance improvements appear to be due to the path lengths needed by the networks to connect distant words in a sentence: linear for RNNs, logarithmic for convolutional models, and constant for the transformer.

In this paper we propose a neural architecture that shares some properties with the above-mentioned models, while maintaining a recurrent design.
Our hypothesis is that current RNNs for NMT have not been designed to take full advantage of deep structures and that better design could lead to improved performance and efficiency.
Contemporary to this work, Chen et al.~\shortcite{chen2018best} have shown that RNN can still outperform the transformer model when using better hyper-parameters.

We start by discussing previous efforts that proposed simplified and theoretically grounded versions of the LSTM RNN, which very recently lead to the so-called Simple Recurrent Unit (SRU). Then, we introduce our NMT architecture based on weakly-recurrent units, which we name Simple Recurrent NMT (SR-NMT).
We present machine translation results on two public benchmark, WMT14 English-German 
and WMT16 English-Romanian, and compare the results of our architecture against LSTM 
and SRU based NMT, using similar settings for all of them. Results show that SR-NMT trains faster than LSTM NMT and outperforms both LSTM and SRU NMT. In particular, SR-NMT with 8-layers even outperforms Google's NMT 8-layer LSTM architecture \cite{wu2016google}. Moreover, training our model took the 
equivalent of $12$ days on a single K80 GPU against the $6$ days on $96$ K80 GPUs reported by \cite{wu2016google}. Finally, the NMT architecture presented in this paper was
developed in OpenNMT-py~\cite{klein2017opennmt} and the code is publicly available on Github\footnote{https://github.com/mattiadg/SR-NMT}.

\section{Related works}
RNNs are an important tool for NMT, and have ranked at the top of the WMT news translation shared tasks~\cite{bojar2017findings} in the last three  years~\cite{luong2015stanford,sennrich2016edinburgh,sennrich2017university}. Recurrent NMT is also the first approach that outperformed phrase-based statistical MT~\cite{bentivogli2016neural}.
Despite the important results, training of RNNs remains inefficient because of an 
intrinsic lack of parallelism and the necessity of redundant parameters in its 
LSTMs and GRUs~\cite{ravanelli2018light,zhou2016minimal} variants.
Sennrich et al.~\shortcite{sennrich2017university} reduce training time in two different ways: by reducing the network size with tied embeddings~\cite{press2017using} and by
adding layer normalization to their architecture~\cite{ba2016layer}. In fact,  the reduction of the covariate shift produced by this mechanism shows to significantly speed up convergence of the training algorithm. Of course, it does not alleviate  the lack of parallelism.

Pascanu et al.~\shortcite{pascanu2014how} studied RNNs and found that the classical stacked RNN architecture does not have a clear notion of \textit{depth}. 
In fact, when performing back-propagation through time, the gradient is sent backward in both the horizontal and vertical dimensions, thus having a double notion of depth, which also hurts the optimization procedure. 
They propose as a solution the notions of \textit{deep transition}, from one hidden state to the following hidden state, and the notion of \textit{deep output}, 
from the last RNN layer to the network output layer.
The winning model in WMT17 actually implemented both of them ~\cite{sennrich2017university,sennrich2017nematus}.

Balduzzi and Ghifary~\shortcite{balduzzi2016strongly} proposed strongly-typed RNNs, which are variants of vanilla RNN, GRU and LSTM that respect some constraints and are theoretically grounded on the concept of \textit{strongly-typed quasi-linear algebra}. 
A strongly-typed quasi-linear algebra imposes constraints on the allowed operations for an RNN. In particular, in this framework there is a constraint inspired from the type system from physics, and one inspired by functional programming. 
The idea of types forbids the sum of vectors generated from different branches of computation. In the case of RNNs, this means that it is not possible to sum among them the previous hidden state and the current input, as they are produced by different computation branches. 
The second constraint aims to simulate the distinction among \textit{pure functions} and functions with side effects, typical of functional programming. 
In fact, as RNNs own a state, they can approximate \textit{algorithms} and also produce ``side effects''.
According to the authors, side effects manifest when the horizontal (time-dimension) connections are altered, and are the reason behind the poor behavior of techniques such as dropout~\cite{srivastava2014dropout} or batch normalization~\cite{ioffe2015batch} when they are applied to the horizontal direction straightforwardly~\cite{laurent2016batch,zaremba2014recurrent}. 
Thus, the side effects should be confined to a part of the network that cannot hinder the learning process.
The solution they propose consists in using learnable parameters only in stateless equations (\textit{learnware}), while the states are combined in parameterless equations (\textit{firmware}).
The combination is achieved through the use of \textit{dynamic average pooling} (or peephole connections), which allows the network to use equations with parameters to compute the states and the gates, and then use the gate vectors to propagate forward horizontally the hidden state.
The authors show theoretically that strongly-typed RNNs have generalization capabilities similar to their classical counterparts, and confirm it with an empirical investigation over several tasks, where the strongly-typed RNNs achieve results not worse than their classical counterparts while training for less time.
In addition, the absence of parameters in the state combination cancels the problem of depth introduced by Pascanu and colleagues, as these models need only the classical back-propagation and not back-propagation through time.

Quasi-recurrent neural networks~\cite{bradbury2017quasi} are an extension of the previous work that use gated convolutions in order to not compute functions of isolated input tokens, but always consider the context given by a convolutional window. 

SRUs~\cite{lei2017sru}, are a development of the units proposed by Balduzzi and Ghifary designed for training speed efficiency. The equations can be easily CUDA optimized, while a good task performance is obtained by stacking many layers in a deep network. SRUs use \textit{highway connections}~\cite{srivastava2015highway} to enable the training of deep networks. 
Moreover, SRUs can parallelize the computation over the time steps also in the decoder. In fact during training the words of the whole sequence are known and there is no dependency on the output of the previous time step.
As for strongly-typed RNNs, the information from the context is propagated with dynamic average pooling, which is much faster to compute than matrix multiplications.
SRUs were tested on a number of tasks, including machine translation, and showed performance similar to LSTMs, but with significantly lower training time. 
However to obtain results comparable to a weak LSTM-based NMT, SRUs require many more layers of computation.
The results show that a single SRU has a significantly lower representation capability than a single LSTM. In addition, every layer adds little overhead in terms of training time per epoch, but also the results show little improvement.

In this work we further develop the idea of SRUs, and propose an NMT architecture that can outperform LSTM-based NMT.

\begin{figure}[t]
\includegraphics[width=2.5\linewidth]{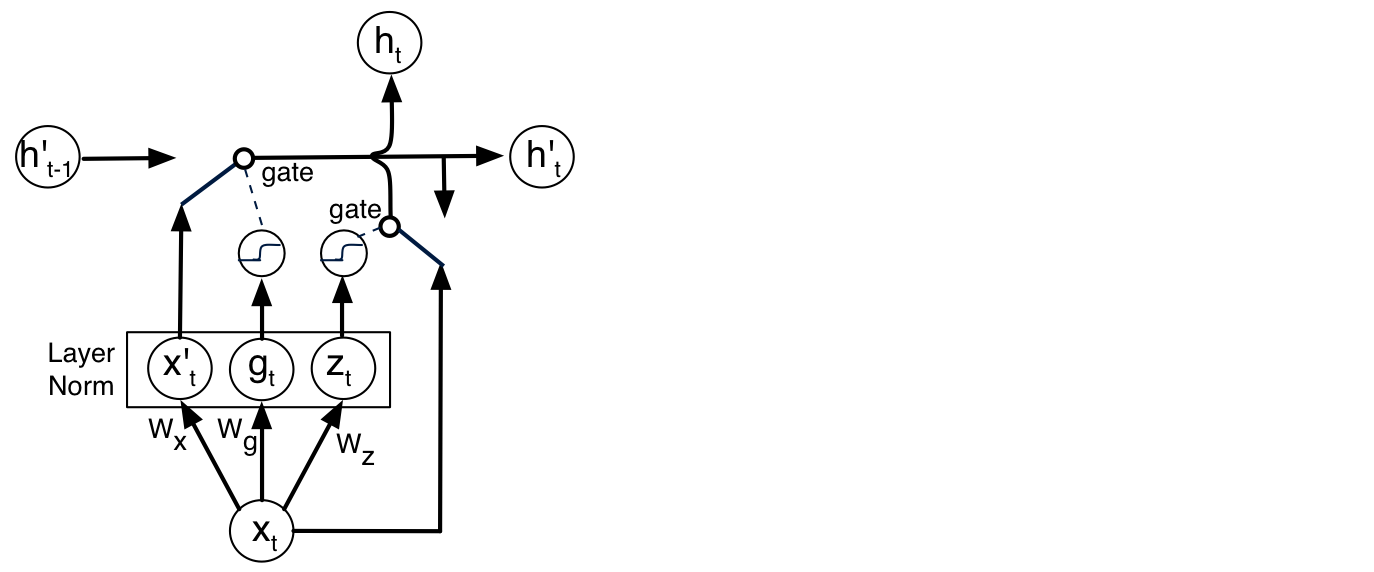}
\caption{Core weakly-recurrent unit used in the SR-NMT architecture. Layer normalization is performed only once for all the transformations. At the end of the unit, the gate $\mathbf{z}_t$ is used for the highway connection.}
\end{figure}

\section{Simple Recurrent NMT}
\begin{figure*}[t]
\includegraphics[width=0.9\textwidth]{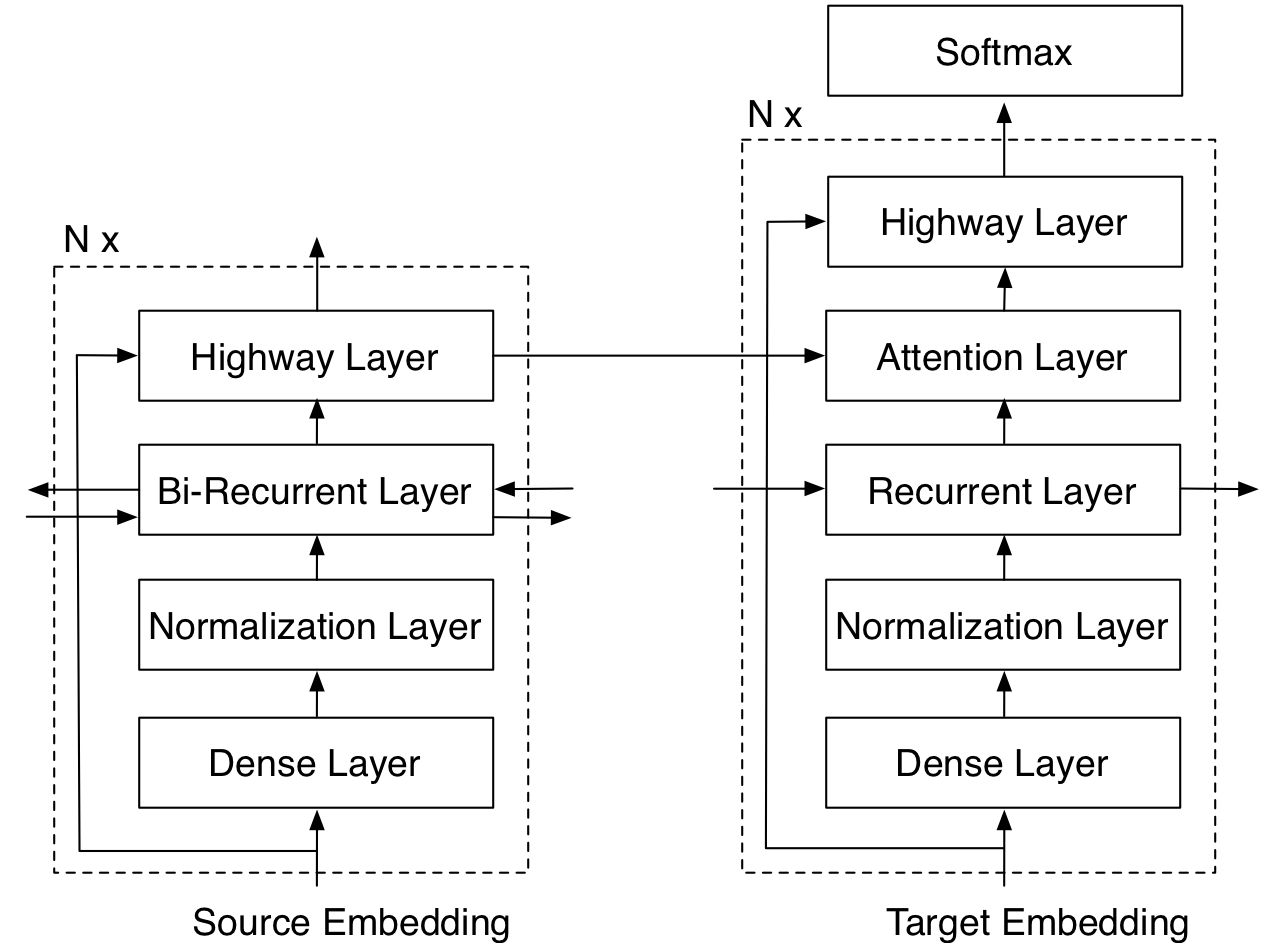}
\caption{SR-NMT encoder-decoder architecture. On the left, a single encoder block to repeat N times. The output of the last layer is used as input for the decoder's attention layers. On the right, a decoder block to repeat N times. The first three sub-layers are the same in the encoder and the decoder, but the latter has an attention layer before the highway connection.}
\label{architecture}
\end{figure*}

We propose a sequence-to-sequence architecture that uses an enhanced version of SRUs (see Figure~1) to improve the training process, in particular with many layers, and increase the representation capability.
In fact, although Lei et al.~\shortcite{lei2017sru} show that they can train networks with up to $10$ layers of SRUs,  both in encoder and decoder, without overfitting, their results are far from the state of the art of recurrent NMT.
Our design goals are addressed in a way similar to ~\cite{gehring2017convolutional,vaswani2017attention}. We add an attention layer within every decoder unit, 
and make the training more stable by adding a layer normalization layer~\cite{ba2016layer} after every matrix multiplication with a parameter matrix.
The layer normalization reduces the covariate shift~\cite{ioffe2015batch}, thus it makes easier the training of deep networks.
In addition to layer normalization, our units use highway connections~\cite{srivastava2015highway}, which enable the training of deep networks.
Our SR-NMT architecture is shown in Figure~\ref{architecture}

Our weakly-recurrent units used in the encoder and decoder both separate \textit{learnware} and \textit{firmware}, although not being strongly typed ~\cite{balduzzi2016strongly} as they include highway connections summing vectors of different types. In the following, we introduce in detail the encoder and decoder networks of our simple recurrent NMT architecture.

\subsection{Encoder}
Our encoder uses bidirectional weakly-recurrent units with layer normalization. We use two candidate hidden states ($\overrightarrow{\mathbf{h}}_i,\overleftarrow{\mathbf{h}}_i$) and two recursion gates ($\overrightarrow{\mathbf{g}}_i,\overleftarrow{\mathbf{g}}_i$) for the two directions. The candidate hidden state for every time step is computed as a weighted average among the current input and the previous hidden state, controlled by the two gates (peephole connections). We apply a single  normalization ({\em LN}) for each layer to improve  training convergence and impose a soft constraint among the parameters. Finally, the input of each layer is combined with its output through highway connections. Formally, our encoder layer is defined by the following equations:\\

\noindent
$\mathbf{x_i}\in R^d;\ \ \  \mathbf{W}\in R^{d\times (4\frac{d}{2}+d)}$\\

\noindent$[\overrightarrow{\mathbf{x}}_i, \overleftarrow{\mathbf{x}}_i, \overrightarrow{\mathbf{g}}_i, \overleftarrow{\mathbf{g}}_i, \mathbf{z_i}]\  =\  LN(\mathbf{x_i W}) $ 
\begin{eqnarray*}
\overrightarrow{\mathbf{h}}_i &=& (1 - \sigma(\overrightarrow{\mathbf{g}}_i)) \odot \overrightarrow{\mathbf{h}}_{i-1} + \sigma(\overrightarrow{\mathbf{g}}_i)) \odot \overrightarrow{\mathbf{x}}_i \\
\overleftarrow{\mathbf{h}}_i &=& (1 - \sigma(\overleftarrow{\mathbf{g}}_i)) \odot \overleftarrow{\mathbf{h}}_{i+1} + \sigma(\overleftarrow{\mathbf{g}}_i) \odot \overleftarrow{\mathbf{x}}_i \\
\mathbf{h}_i &=& (1 - \sigma(\mathbf{z}_i)) \odot [\overrightarrow{\mathbf{h}}_i; \overleftarrow{\mathbf{h}}_i] + \sigma(\mathbf{z}_i) \odot \mathbf{x}_i
\end{eqnarray*}

\begin{table}[t]
\centering
\begin{tabular}{l|l}
\textbf{Model} & \textbf{train speed}  \\ 
\specialrule{.1em}{.05em}{.05em} 
LSTM 2L            & 3700 tok/s\\ \hline
SRU 3L             & 4600 tok/s\\ \hline
SR-NMT 1L        & 7900 tok/s\\
SR-NMT 2L        & 5500 tok/s\\
SR-NMT 3L        & 4300 tok/s\\
SR-NMT 4L        & 3600 tok/s\\
\specialrule{.1em}{.05em}{.05em}
   
\end{tabular}
\caption{Training speed comparison of our architectures with LSTM and SRU baselines on WMT14 En-De. Timings are performed on a single Nvidia Gtx 1080 GPU with CUDA 8.0 and pytorch 0.2.}
\label{speed}
\end{table}

\subsection{Decoder}
The decoder employs unidirectional units, with layer normalization ({\em LN}) after every matrix multiplication similarly to the encoder units, and has an attention mechanism in every layer. 
The attention output is combined with the layer's hidden state in a way similar to the \textit{deep output}~\cite{pascanu2014how} used by Luong~\shortcite{luong2015stanford}. The highway connection is applied only at the end of the unit.
The presence of multiple attention models connected to the last encoder layer produces a high gradient for the encoder output, thus we scale the gradient dividing the attention output by $\sqrt{d}$.
This kind of scaling has been proposed in \cite{vaswani2017attention} inside the transformer model, but we observed empirically that this version works better for our model. Formally:\\

\noindent$\mathbf{y_i}\in R^d;\ \ \mathbf{W}\in R^{d \times 3d};\ \  \mathbf{W_s, W_c}\in R^{d\times d} $

\begin{eqnarray*}
[\tilde{\mathbf{y_i}}, \mathbf{g_i}, \mathbf{z_i}] &=& LN(\mathbf{y_i W})\\
\mathbf{\tilde{s_i}} &=& (1 - \sigma(\mathbf{g_i})) \odot \mathbf{\tilde{s}_{i-1}} + \sigma(\mathbf{g_i}) \odot \tilde{\mathbf{y_i}}\\
\mathbf{c_i} &=& attn(\mathbf{\tilde{s_i}}, \mathbf{H}) (1 / \sqrt[]{d}) \\
\mathbf{o_i} &=& \tanh(LN(\mathbf{\tilde{s_i} W_s}) + LN( \mathbf{c_i W_c})) \\
\mathbf{s_i} &=& (1 - \sigma(\mathbf{z_i})) \odot \mathbf{o_i} + \sigma(\mathbf{z_i}) \odot \mathbf{y_i}
\end{eqnarray*}
The decoder includes a standard {\em softmax} layer over the target vocabulary which is omitted from this description.  For our architecture, we opted for a layer-normalized version of the MLP global attention \cite{bahdanau2014neural}, which showed to perform better than the dot attention model \cite{luong2015effective}:
\begin{eqnarray*}
\mathbf{\tilde{\alpha_{ij}}} &=& \mathbf{v_\alpha}\tanh(LN(\mathbf{\tilde{s_i} W_{as}}) + LN(\mathbf{h_j W_{ah}})) \\
\boldsymbol{\alpha_i} &=& \textnormal{softmax}(\tilde{\boldsymbol{\alpha}}_{i}) \\
\mathbf{c_i} &=& \sum_{i=0}^L \alpha_{ij} \mathbf{h_j}
\end{eqnarray*}

\noindent
Our SR-NMT architecture stacks several layers both on the encoder and   decoder sides, as shown in Figure~2. The natural structure we consider is  one having the same number of layers on both sides, although different topologies could be considered, too. 

\section{Experiments}

We implemented our architecture in PyTorch~\cite{paszke2017automatic} inside the OpenNMT-py toolkit~\cite{klein2017opennmt}.
All the tested models have been trained with the Adam~\cite{kingma2015adam} optimizer until convergence, using the typical initial learning rate of $0.0003$, and default values for $\beta_1$ and $\beta_2$. 
At convergence, the models were further trained until new convergence with learning rate $0.00015$~\cite{bahar2017empirical}.
The model used to restart the training is selected according to the perplexity on the validation set.
We applied dropout of $0.1$ before every multiplication by a parameter matrix, and in the case of LSTM it is applied only to vertical connections in order to use the LSTM version optimized in CUDA. 
The batch size is $64$ for all the experiments and all the layers for all the models have an output size of $500$. 

\subsection{Datasets}
We used as benchmarks the WMT14 English to German and the WMT16 English to Romanian datasets. 

In the case of WMT14 En-De, the training set consists of the concatenation of all the training data that were available for the 2014 shared task, the validation set is the concatenation of newstest2012 and 2013, and newstest2014 is our test set. 
Then, it was preprocessed with tokenization, punctuation normalization and de-escaping of the special characters.
Furthermore, we applied BPE segmentation~\cite{sennrich2016neural} with 32,000 merge rules.
We removed from the training data all the sentence pairs where the length of at least one sentence exceeded 50 tokens, resulting on a training set of $3.9M$ sentence pairs. 
Furthermore, we cleaned the training set by removing sentences in a wrong language and poorly aligned sentence pairs. For the cleaning process we used the automatic pipeline developed by the ModernMT project\footnote{https://github.com/ModernMT/MMT}.

In the case of WMT16 En-Ro, we have used the same data and preprocessing used by Sennrich et al.~\shortcite{sennrich2016edinburgh} and Gehring et al.~\shortcite{gehring2017convolutional}. The back-translations to replicate the experiments are available\footnote{http://data.statmt.org/rsennrich/wmt16\_backtranslations/en-ro
.} and we applied the same preprocessing\footnote{https://github.com/rsennrich/wmt16-scripts/blob/80e21e5/
sample/preprocess.sh}, which involves punctuation normalization, tokenization, truecasing and BPE with $40K$ merge rules. 


   

\section{Evaluation}
In this section, we describe the evaluation of our models with the two benchmarks. As our main goal is to prove that SR-NMT represent a valid alternative to LSTMs, we have put more effort on WMT14 En-De, which is widely used as a benchmark dataset. The experiments on WMT16 En-Ro are aimed to verify the effectiveness of our models in a different language pair with a different data size.

\subsection{WMT14 English to German}
The results for WMT14 En-De are evaluated on cased output, tokenized with the tokenizer script from the Moses toolkit~\cite{koehn2007moses}, and the BLEU score is computed using \textnormal{multi-bleu.pl} from the same toolkit. With this procedure the results are comparable with the results reported from the other publications\footnote{https://github.com/tensorflow/tensor2tensor/blob/master/
tensor2tensor/utils/get\_ende\_bleu.sh}. 

We compare our models with the results reported in~\cite{lei2017sru}, and also reproduce some of their experiments. We train $4$ baseline models following the same procedure used for SR-NMT. Three baselines are LSTM-based NMT models as provided by OpenNMT-py, with $2$, $3$ and $5$ layers in both encoder and decoder. The other is an SRU model with $3$ layers that we re-implemented in PyTorch, in order to perform a more fair comparison with our model.
For the baselines we use dropout after every layer and MLP attention~\cite{luong2015effective}, both resulting in better results than the default implementation.
Furthermore, we compare our results with Google's NMT system~\cite{wu2016google}, Convolutional S2S model ~\cite{gehring2017convolutional}, and the Transformer~\cite{vaswani2017attention}.

\begin{table}[t]
\centering
\begin{tabular}{l|l|l}
\hline
\textbf{WMT14 En-De} & \textbf{BLEU}  & \textbf{\# par} \\ 
\specialrule{.1em}{.05em}{.05em} 
LSTM 2L            & 21.82  &  62M \\
LSTM 3L            & 22.26  &  65M \\
LSTM 5L            & 22.72  &  72M \\ \hline
SRU 3L             & 20.88  &  59M \\ \hline
SR-NMT 1L        & 18.33  & 56M  \\
SR-NMT 2L        & 21.82  & 58M \\
SR-NMT 3L        & 22.35  & 61M \\
SR-NMT 4L        & 23.32  & 63M \\
SR-NMT 5L        & 24.11  & 66M \\
SR-NMT 6L        & 23.93  & 68M \\
SR-NMT 7L        & 24.34  & 71M \\
SR-NMT 8L        & 24.87  & 73M \\
SR-NMT 9L        & 25.04  & 76M \\
SR-NMT 10L       & 24.98  & 78M \\

\specialrule{.1em}{.05em}{.05em} 
\multicolumn{3}{l}{\textbf{Setting of ~\cite{lei2017sru}}} \\ \hline
LSTM 2L    & 19.67 & 84M  \\ 
LSTM 5L    & 20.45 & 96M  \\ 
SRU 3L     & 18.89 & 81M  \\
SRU 10L    & 20.70 & 91M  \\

\specialrule{.1em}{.05em}{.05em} 
\multicolumn{3}{l}{\textbf{GNMT} \cite{wu2016google}} \\ \hline
LSMT 8L     & 24.61 & -  \\
Ensemble & 26.30 & -  \\
\specialrule{.1em}{.05em}{.05em} 
\multicolumn{3}{l}{\textbf{Convolutional} \cite{gehring2017convolutional}} \\ \hline
ConvS2S 15L   & 25.16 & -  \\
Ensemble  & 26.43 & -  \\

\specialrule{.1em}{.05em}{.05em} 
\multicolumn{3}{l}{\textbf{Transformer} \cite{vaswani2017attention}} \\ \hline
Base 6L    & 27.30 & 65M \\ 
Big 6L & \textbf{28.40} & 213M  \\ 
\specialrule{.1em}{.05em}{.05em} 
\end{tabular}
\caption{Experiments with cleaned data on WMT14 En-De both for our architectures and the baselines, and comparison with the state of the art.}
\label{clean}
\end{table}

\subsection{WMT16 English to Romanian}
In the case of English to Romanian, we trained our models with the same hyper-parameters used for English to German, despite the difference in the amount of data.
The BLEU score is computed using the official script of the shared task\footnote{mteval-v13a.pl}, which runs on cased and detokenized output. 

We did not implement baselines for this language pair, and we compare our results with the winning submission of the WMT16 shared task~\cite{sennrich2016edinburgh}, with the Convolutional S2S model~\cite{gehring2017convolutional} and the Transformer~\cite{gu2018nonautoregressive}.

\begin{table}[t]
\centering
\begin{tabular}{l|l}
\hline
\textbf{WMT16 En-Ro} & \textbf{BLEU}  \\ 
\specialrule{.1em}{.05em}{.05em} 
SR-NMT 1L        & 24.74 \\
SR-NMT 2L        & 26.41 \\
SR-NMT 4L        & 28.81 \\
SR-NMT 6L        & 29.04 \\
SR-NMT 8L        & 28.70 \\
\specialrule{.1em}{.05em}{.05em} 
\multicolumn{2}{l}{\textbf{GRU} \cite{sennrich2016edinburgh}} \\ \hline
GRU 1L+2L    & 28.1  \\
Ensemble & 28.2  \\
\specialrule{.1em}{.05em}{.05em} 
\multicolumn{2}{l}{\textbf{Convolutional} \cite{gehring2017convolutional}} \\ \hline
ConvS2S 15L   & 30.02 \\
\specialrule{.1em}{.05em}{.05em} 
\multicolumn{2}{l}{\textbf{Transformer} \cite{gu2018nonautoregressive}} \\ \hline
NAT     & 29.79 \\ 
Transformer  & \textbf{31.91}  \\ 
\specialrule{.1em}{.05em}{.05em} 
\end{tabular}
\caption{Results on the test set of WMT16 En-Ro and comparison with the state of the art.}
\label{enro}
\end{table}

\section{Results}
In this section, we discuss the performance in terms of training speed and translation quality of our architecture. 

\subsection{WMT14 En-De}
In the first part of Table~\ref{clean} we list the results of SR-NMT using from $1$ up to $10$ layers and our baselines.
The training speeds are reported in Table~\ref{speed}.

SR-NMT with $3$ layers has a number of parameters comparable to the LSTM baseline with $2$ layers, but its training speed is $14\%$ faster (4300 tok/s vs 3700 tok/s), and the BLEU score is $0.5$ points higher.
Moreover, the implementation of the LSTM is optimized at CUDA level, while our architecture is fully implemented in PyTorch and could be made faster following the optimizations of Lei et al.~\shortcite{lei2017sru}.
Furthermore, also the layer normalization can be implemented faster in CUDA\footnote{https://github.com/MycChiu/fast-LayerNorm-TF}.
By increasing the number of LSTM layers from $2$ to $5$, the improvement in terms of BLEU score is only $0.9$ points, and it is worse than SR-NMT with $4$ layers.

\begin{figure}[t]
\includegraphics[width=\linewidth]{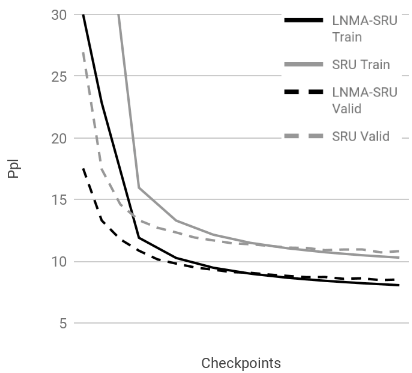}
\caption{Perplexity against time for SR-NMT and SRU-based NMT with 3 layers and the same optimization policy. The convergence is achieved after a comparable number of iterations, but SR-NMT achieves a better convergence point.}
\label{curve}
\end{figure}

The comparison with NMT based on SRUs is in favor of our architecture, which achieves higher translation quality with less layers. In particular, SR-NMT with $2$ layers outperforms SRU NMT with $3$ layers by $1$ BLEU point and also trains faster (Table~\ref{speed}). However, this comparison is performed with implementations that are not optimized for fast execution in GPU. A speed comparison with optimized implementations could lead to different results. 

In the second part of Table~\ref{clean} we report some results from~\cite{lei2017sru} on the same benchmark. The different number of parameters is probably due to a different size of the vocabulary, in fact the number of merge rules used is not reported in the paper. Our LSTM baseline performs clearly better then the one cited because of the straightforward improvements we implemented, i.e. the use of \textit{input feeding}~\cite{luong2015effective}, MLP attention instead of general or dot attention, and dropout in every layer.
With this improvements, our baseline with $2$ layers obtains $1.4$ BLEU scores more than its counterpart using $5$ layers. Moreover, it also outperforms SRU with $10$ layers by more than $1$ point. This result shows that our additions are fundamental to have a competitive  architecture based on weakly recurrent units.

\noindent
Figure~\ref{curve} shows a comparison of the learning curves of SR-NMT and SRU NMT both with $3$ layers. We can easily observe that the convergence of SR-NMT occurs at comparable speed but to a better point and the validation perplexities of the two models are very close to the training perplexities. 

\noindent
When we compare SR-NMT to GNMT (Table~2), we can observe that SR-NMT with $8$ layers performs slightly better than GNMT, which in turn uses many more parameters, as it uses 8 LSTM layers with size $1024$. Moreover, GNMT was trained for $6$ days on $96$ Nvidia K80 GPUs, while our model took the equivalent\footnote{As our training currently works on single GPU, we could only fit models up to 7 layers into a K80, hence the estimate. Actually, models above  7 layers were trained on a V100 GPU.} of $12$ days on a single K80 GPU.  

Our best BLEU score, 25.04, is obtained with $9$ layers. This is only $0.12$ BLEU points below the convolutional model that used $15$ layers in both encoder and decoder, and hidden sizes of at least $512$. Finally, we notice that SR-NMT's best performance is still below that of the transformer model. Future work will be devoted to deeper explore the hyper-parameter space of our architecture and   enhance it along the recent developments in \cite{chen2018best}. 

\subsection{WMT16 En-Ro}
The results for WMT16 En-Ro are listed in Table~\ref{enro}. We obtain the highest score for this dataset with $6$ layers, which can be due to the smaller dimension of the dataset, for which we did not add any form of regularization.

Our best SR-NMT system, which obtained a BLEU score of $29.04$, is $1$ BLEU point lower than ConvS2S, and almost $3$ BLEU points lower than the state of the art.
Nonetheless, this score is almost $1$ BLEU point better than the score obtained by the winning system in WMT16~\cite{bojar2016findings}, showing that SR-NMT represent a viable alternative to more complex RNNs.

\section{Ablation experiments}
In this section, we evaluate the importance of our enhancements to the original SRU unit, namely multi-attention and layer normalization, and  of the highway connections, which were already present in the original formulation of SRUs.

We take our SR-NMT model with $4$ layers and remove from it one component or a set of components. All the combinations are reported. Results refer to the WMT14 En-De task after performing only one training stage. In other words, we did not restart training after convergence as we did for the systems reported in Table~2. As our previous experiments already proved the superiority of SR-NMT to LSTMs, the goal of this section is to understand whether all the proposed additions are important and to quantify their contributions.

\begin{table}[t]
\centering
\begin{tabular}{l|l|l}
\hline
\textbf{Model} & \textbf{BLEU} & \textbf{$\Delta$} \\ 
\specialrule{.1em}{.05em}{.05em} 
SR-NMT 4L        & 22.99 & 0 \\ \hline
- LayerNorm        & 21.97 & -1.02 \\
- Multi Attention  & 21.57 & -1.42 \\
- Highway          & 20.85 & -2.14 \\
- Ln \& MA         & 20.51  & -2.48 \\
- LN \& highway    & / & - / \\
- MA \& highway    &  19.54 & -3.45\\
- LN, MA \& highway & 18.39 & -4.6 \\
\specialrule{.1em}{.05em}{.05em} 
\end{tabular}
\caption{Ablation experiments on SR-NMT with $4$ layers. BLEU scores are computed after one training stage. While removing multi attention we still keep one attention model in the last layer. The system without layer normalization and highway connections failed to converge.}
\label{ablation}
\end{table}

From Table~\ref{ablation} we can observe that the removal of highway connections causes the highest drop in performance ($-2.14$ BLEU points), followed by multi attention and then layer normalization. 
Another important observation is the additivity of the contributions from all the components, in fact when two or three components are removed at once, the drop in performance is roughly the sum of the drops caused by the single components.
Finally, the removal of layer normalization and highway connections, while keeping multi attention, causes a gradient explosion that prevents the SR-NMT system from converging.

\section{Conclusions}
In this paper we have presented a simple recurrent NMT architecture that enhances previous SRUs~\cite{lei2017sru} by adding elements of other architectures, namely layer normalization and multiple attentions.
Our goal was to explore the possibility to make weakly-recurrent units competitive with LSTMs for NMT. 
We have shown that our SR-NMT architecture is able to outperform more complex LSTM NMT models on two public benchmarks. In particular, SR-NMT performed even better than the GNMT system, while using a simpler optimization policy, a vanilla beam search and a fraction of its computational resources for training. Future work will be in the direction of further enhancing SR-NMT by integrating core components 
that seem to particularly boost performance of the best non recurrent NMT architectures. 


\section*{Acknowledgements}
We would like to thank Nicola Bertoldi for his technical support and the anonymous reviewers for their useful comments. 
We gratefully acknowledge the support of NVIDIA Corporation with the donation of GPUs used for this research.



\begin{thebibliography}{}

\bibitem[\protect\citename{Ba et al.}2016]{ba2016layer}
Ba, Jimmy Lei, Jamie Ryan Kiros Geoffrey E. Hinton.
\newblock 2016.
\newblock Layer normalization.
\newblock arXiv preprint arXiv:1607.06450.

\bibitem[\protect\citename{Bahar et al.}2017]{bahar2017empirical}
Bahar, Parnia, Tamer Alkhouli, Jan-Thorsten Peter, Christopher J. S. Brix, and Hermann Ney.
\newblock 2017.
\newblock Empirical investigation of optimization algorithms in neural machine translation. \newblock {\em The Prague Bulletin of Mathematical Linguistics, 108(1)}, 13-25.  
  
\bibitem[\protect\citename{Bahdanau et al.}2015]{bahdanau2014neural}
Bahdanau, Dzmitry, Kyunghyun Cho and Yoshua Bengio.
\newblock 2015.
\newblock Neural machine translation by jointly learning to align and translate 
\newblock {\em Proceedings of the 3rd International Conference on Learning Representations,} San Diego, USA.

\bibitem[\protect\citename{Balduzzi and Ghifary}2016]{balduzzi2016strongly}
Balduzzi, David and Muhammad Ghifary.
\newblock 2016.
\newblock Strongly-Typed Recurrent Neural Networks.
\newblock {\em International Conference on Machine Learning}, 1292--1300.

\bibitem[\protect\citename{Barone et al.}2017]{barone2017deep}
Barone, Antonio Valerio Miceli, Jindrich Helcl, Rico Sennrich, Barry Haddow and Alexandra Birch.
\newblock 2017.
\newblock Deep architectures for Neural Machine Translation.
\newblock {\em Proceedings of the Second Conference on Machine Translation}, Copenhagen, Denmark. 99-107.  


\bibitem[\protect\citename{Bentivogli et al.}2016]{bentivogli2016neural}
Bentivogli, Luisa, Arianna Bisazza, Mauro Cettolo, Marcello Federico.
\newblock 2016.
\newblock Neural versus Phrase-Based Machine Translation Quality: a Case Study.
\newblock {\em Conference on Empirical Methods in Natural Language Processing (EMNLP)}, November 1-5, 2016, Austin, Texas, USA.
\newblock 257--267
\bibitem[\protect\citename{Bradbury et al.}2017]{bradbury2017quasi}
Bradbury, James, Stephen Merity, Caiming Xiong and Richard Socher.
\newblock 2017.
\newblock Quasi-Recurrent Neural Networks.
\newblock {\em Proceedings of the 5th International Conference on Learning Representations}, Toulon, France. 

\bibitem[\protect\citename{Bojar et al.}2016]{bojar2016findings}
Bojar,  Ond{\v{r}}ej, Rajen Chatterjee, Christian Federmann, Yvette Graham, Barry Haddow, Matthias Huck, Antonio Jimeno Yepes, Philipp Koehn, Varvara Logacheva, Christof Monz, Matteo Negri, Aurelie Neveol, Mariana Neves, Martin Popel, Matt Post, Raphael Rubino, Carolina Scarton, Lucia Specia, Marco Turchi, Karin Verspoor, Marcos Zampieri.
\newblock 2016.
\newblock Findings of the 2016 conference on machine translation (wmt16).
\newblock {\em Proceedings of the First Conference on Machine Translation}, Berlin, Germany. 131--198.

\bibitem[\protect\citename{Bojar et al.}2017]{bojar2017findings}
Bojar, Ond{\v{r}}ej, Rajen Chatterjee, Christian Federmann, Yvette Graham, Barry Haddow, Shujian Huang, Matthias Huck, Philipp Koehn, Qun Liu, Varvara Logacheva, Christof Monz, Matteo Negri, Matt Post, Raphael Rubino, Lucia Specia, Marco Turchi.
\newblock 2017.
\newblock Findings of the 2017 conference on machine translation (wmt17).
\newblock {\em Proceedings of the Second Conference on Machine Translation}, Copenhagen, Denmark. 169--214.

\bibitem[\protect\citename{Cettolo et al.}2012]{cettolo2012wit}
Cettolo, Mauro, Christian Girardi and Marcello Federico.
\newblock 2012.
\newblock Wit3: Web inventory of transcribed and translated talks.
\newblock {\em Proceedings of the 16th Annual Conference of the European Association for Machine Translation}, Trento, Italy.

\bibitem[\protect\citename{Chen et al.}2018]{chen2018best}
Chen, Mia Xu, et al.
\newblock 2018.
\newblock The Best of Both Worlds: Combining Recent Advances in Neural Machine Translation. 
\newblock {\em arXiv preprint arXiv:1804.09849.}

\bibitem[\protect\citename{Cho et al.}2014]{cho2014properties}
Cho, Kyunghyun, Bart Van Merri{\"e}nboer, Dzmitry Bahdanau and Yoshua Bengio.
\newblock 2014.
\newblock On the properties of neural machine
translation: Encoder-decoder approaches
\newblock {\em Syntax, Semantics and Structure in Statistical Translation (2014): 103.} 

\bibitem[\protect\citename{Gal et al.}2016]{gal2016theoretically}
Gal, Yarin and Zoubin Ghahramani.
\newblock 2016.
\newblock A theoretically grounded application of dropout in recurrent neural networks. 
\newblock {\em In Advances in neural information processing systems}, 1019--1027.

\bibitem[\protect\citename{Gehring et al.}2017]{gehring2017convolutional}
Gehring, Jonas, Michael Auli, David Grangier, Denis Yarats and Yann N. Dauphin. 
\newblock 2017.
\newblock Convolutional Sequence to Sequence Learning. 
\newblock {\em Proceedings of the 34th International Conference on Machine Learning}, Sidney, Australia. 1243--1252.

\bibitem[\protect\citename{Gu et al.}2018]{gu2018nonautoregressive}
Gu, Jiatao, James Bradbury, Caiming Xiong, Victor O.K. Li, Richard Socher.
\newblock 2018. 
\newblock Non-Autoregressive Neural Machine Translation. 
\newblock {\em Proceedings of the Sixth International Conference on Learning Representations (ICLR)}, Vancouver, Canada.


\bibitem[\protect\citename{Hochreiter and Schmidhuber}1997]{hochreiter1997long}
Hochreiter, Sepp and J{\"u}rgen Schmidhuber. 
\newblock 1997.
\newblock Long short-term memory
\newblock {\em Neural computation.} 
\newblock MIT Press
\newblock 1735--1780.


\bibitem[\protect\citename{Ioffe and Szegedy}2015]{ioffe2015batch}
Ioffe, Sergey and Christian Szegedy.
\newblock 2015.
\newblock Batch normalization: Accelerating deep network training by reducing internal covariate shift.
\newblock {\em International conference on machine learning}, 448--456.


\bibitem[\protect\citename{Kalchbrenner et al.}2016]{Bytenet}
Kalchbrenner, Nal, Lasse Espeholt, Karen Simonyan, Aaron van den Oord, Alex Graves and Koray Kavukcuoglu.
\newblock 2016
\newblock Neural machine translation in linear time.
\newblock {\em arXiv preprint arXiv:1610.10099.}

\bibitem[\protect\citename{Kingma and Ba}2015]{kingma2015adam}
Kingma, Diederik P. and Jimmy Ba.
\newblock 2015.
\newblock Adam: A method for stochastic optimization.
\newblock {\em 3rd International Conference for Learning Representations}, San Diego, USA.

\bibitem[\protect\citename{Klein et al.}2017]{klein2017opennmt}
Klein, Guillaume, Yoon Kim, Yuntian Deng, Jean Senellart and
Alexander M. Rush.
\newblock 2017.  
\newblock {\em OpenNMT: Open-Source Toolkit for Neural Machine Translation.}
\newblock {\em Proceedings of the 55th Annual Meeting of the Association for Computational Linguistics (ACL)}, System Demonstrations, 67-72.

\bibitem[\protect\citename{Koehn et al.}2007]{koehn2007moses}
Koehn, Philipp, Hieu Hoang, Alexandra Birch, Chris Callison-Burch, Marcello Federico, Nicola Bertoldi,  	Brooke Cowan, Wade Shen, Christine Moran, Richard Zens, Chris Dyer, Ondřej Bojar, Alexandra Constantin, Evan Herbst.
\newblock 2007.
\newblock Moses: Open source toolkit for statistical machine translation.
\newblock {\em Proceedings of ACL on interactive poster and demonstration sessions}, 177--180.

\bibitem[\protect\citename{Laurent et al.}2016]{laurent2016batch}
Laurent, C{\'e}sar, Gabriel Pereyra,  Phil{\'e}mon Brakel, Ying Zhang and Yoshua Bengio.
\newblock 2016.
\newblock Batch normalized recurrent neural networks.
\newblock {\em IEEE International Conference on Acoustics, Speech and Signal Processing (ICASSP)}. 2657--2661.
  
\bibitem[\protect\citename{Lei et al.}2017a]{lei2017deriving}
Lei, Tao and Wengong Jin, Regina Barzilay and Tommi Jaakkola.
\newblock 2017.
\newblock Deriving Neural Architectures from Sequence and Graph Kernels.
\newblock {\em Proceedings of the 34th International Conference on Machine Learning}, Sidney, Australia. 2024--2033.

\bibitem[\protect\citename{Lei et al.}2017b]{lei2017sru}
Lei, Tao, Yu Zhang and Yoav Artzi.
\newblock 2017.
\newblock Training RNNs as Fast as CNNs.
\newblock {\em arXiv preprint arXiv:1709.02755}.

\bibitem[\protect\citename{Luong et al.}2015]{luong2015effective}
Luong, Thang, Hieu Pham, and Christopher D. Manning. 
\newblock 2015.
\newblock Effective Approaches to Attention-based Neural Machine Translation.
\newblock {\em Proceedings of the 2015 Conference on Empirical Methods in Natural Language Processing.}
\newblock 1412--1421.

\bibitem[\protect\citename{Luong and Manning}2015]{luong2015stanford}
Luong, Minh-Thang and Christopher D. Manning.
\newblock 2015.
\newblock {Stanford neural machine translation systems for spoken language domains}.
\newblock {\em Proceedings of the 12th International Workshop on Spoken Language Translation}, Da Nang, Vietnam. 76--79.

\bibitem[\protect\citename{Papineni et al.}2002]{papineni2002bleu}
Papineni, Kishore, Salim Roukos, Todd Ward and Wei-Jing Zhu.
\newblock 2002.
\newblock BLEU: a Method for Automatic Evaluation of Machine Translation
\newblock {\em Proceedings of the 40th Annual Meeting of the Association for Computational Linguistics (ACL).}
\newblock 311--318.

\bibitem[\protect\citename{Pascanu et al.}2013]{pascanu2013difficulty}
Pascanu, Razvan, Tomas Mikolov and Yoshua Bengio. 
\newblock 2013.
\newblock On the difficulty of training recurrent neural networks.
\newblock {\em Proceedings of the 30th International Conference on Machine Learning}, Atlanta, USA. 1310--1318.

\bibitem[\protect\citename{Pascanu et al.}2014]{pascanu2014how}
Pascanu, Razvan, Caglar Gulcehre, Kyunghyun Cho and Yoshua Bengio.
\newblock 2014.
\newblock How to construct deep recurrent neural networks.
\newblock {\em Proceedings of the 2nd International Conference on Learning Representations}, Banff, Canada. 

\bibitem[\protect\citename{Paszke et al.}2017]{paszke2017automatic}
Paszke, Adam, Sam Gross, Soumith Chintala, Gregory Chanan, Edward Yang, Zachary DeVito, Zeming Lin, Alban Desmaison, Luca Antiga and Adam Lerer.
\newblock 2017.
\newblock Automatic differentiation in PyTorch.
\newblock {\em NIPS 2017 Autodiff Workshop}, Long Beach, USA.

\bibitem[\protect\citename{Press and Wolf}2017]{press2017using}
Press, Ofir and Lior Wolf.
\newblock 2017.
\newblock Using the Output Embedding to Improve Language Models.
\newblock {\em Proceedings of the 15th Conference of the European Chapter of the Association for Computational Linguistics}: Volume 2, Short Papers (Vol. 2, pp. 157-163).

\bibitem[\protect\citename{Ravanelli et al.}2018]{ravanelli2018light}
Ravanelli, Mirco, Philemon Brakel, Maurizio Omologo, and Yoshua Bengio.
\newblock 2018.
\newblock Light Gated Recurrent Units for Speech Recognition.
\newblock {\em IEEE Transactions on Emerging Topics in Computational Intelligence 2}, no. 2 (2018): 92-102.

\bibitem[\protect\citename{Sennrich et al.}2016a]{sennrich2016neural}
Sennrich, Rico, Barry Haddow, and Alexandra Birch. 
\newblock 2016.
\newblock Neural Machine Translation of Rare Words with Subword Units.
\newblock {\em Proceedings of the 54th Annual Meeting of the Association for Computational Linguistics}, Berlin, Germany.
\newblock 1715--1725.

\bibitem[\protect\citename{Sennrich et al.}2016b]{sennrich2016edinburgh}
Sennrich, Rico, Barry Haddow and Alexandra Birch.
\newblock 2016.
\newblock Edinburgh Neural Machine Translation Systems for WMT 16.
\newblock {\em Proceedings of the First Conference on Machine Translation: Volume 2, Shared Task Papers}, Vol. 2, 371--376.

\bibitem[\protect\citename{Sennrich et al.}2017a]{sennrich2017nematus}
Sennrich, Rico, Orhan Firat, Kyunghyun Cho, Alexandra Birch, Barry Haddow, Julian Hitschler, Marcin Junczys-Dowmunt, Samuel Läubli, Antonio Valerio Miceli Barone, Jozef Mokry, Maria Nădejde.
\newblock 2017.
\newblock Nematus: a Toolkit for Neural Machine Translation.
\newblock {\em Proceedings of the Software Demonstrations of the 15th Conference of the European Chapter of the Association for Computational Linguistics}, Valencia, Spain. 65--68.

\bibitem[\protect\citename{Sennrich et al.}2017b]{sennrich2017university}
Sennrich, Rico, Alexandra Birch, Anna Currey, Ulrich Germann, Barry Haddow, Kenneth Heafield, Antonio Valerio Miceli Barone and Philip Williams.
\newblock 2017.
\newblock {The University of Edinburgh's Neural MT Systems for WMT17}.
\newblock {\em Proceedings of the Second Conference on Machine Translation}, Copenhagen, Denmark. 389--399.

\bibitem[\protect\citename{Siegelmann and Sontag}1995]{siegelmann1995computational}
Siegelmann, Hava T. and Eduardo D. Sontag.
\newblock 1995.
\newblock On the computational power of neural nets. 
\newblock {\em Journal of computer and system sciences}, 50(1), 132-150.

\bibitem[\protect\citename{Srivastava et al.}2015]{srivastava2015highway}
Srivastava, Rupesh Kumar, Klaus Greff, and J{\"u}rgen Schmidhuber. 
\newblock 2015.
\newblock Highway networks.
\newblock ICML 2015 Deep Learning Workshop.
  
\bibitem[\protect\citename{Srivastava et al.}2014]{srivastava2014dropout} 
Srivastava, Nitish, Geoffrey Hinton, Alex Krizhevsky, Ilya Sutskever and Ruslan Salakhutdinov.
\newblock 2014
\newblock  Dropout: A simple way to prevent neural networks from overfitting.
\newblock {\em The Journal of Machine Learning Research}, Vol. 15,
n. 1, 1929--1958.

\bibitem[\protect\citename{Sutskever et al.}2014]{sutskever2014sequence}
Sutskever, Ilya, Oriol Vinyals and Quoc V. Le. 
\newblock 2014.
\newblock Sequence to sequence learning with neural networks.
\newblock {\em Advances in neural information processing systems.}
\newblock 3104--3112.

\bibitem[\protect\citename{Vaswani et al.}2017]{vaswani2017attention}
Vaswani, Ashish, Noam Shazeer, Niki Parmar, Jacob Uszkoreit, Llion Jones, Aidan N. Gomez, Lukasz Kaiser and Illia Polosukhin.
\newblock 2017. 
\newblock Attention is all you need. 
In Advances in Neural Information Processing Systems (pp. 6000-6010).

\bibitem[\protect\citename{Wu et al.}2016]{wu2016google}
Wu, Yonghui,Mike Schuster, Zhifeng Chen, Quoc V. Le, Mohammad Norouzi, Wolfgang Macherey, Maxim Krikun, Yuan Cao, Qin Gao, Klaus Macherey, Jeff Klingner, Apurva Shah, Melvin Johnson, Xiaobing Liu, Łukasz Kaiser, Stephan Gouws, Yoshikiyo Kato, Taku Kudo, Hideto Kazawa, Keith Stevens, George Kurian, Nishant Patil, Wei Wang, Cliff Young, Jason Smith, Jason Riesa, Alex Rudnick, Oriol Vinyals, Greg Corrado, Macduff Hughes and Jeffrey Dean.
\newblock 2016.
\newblock Google’s Neural Machine Translation System: Bridging the Gap between Human and Machine Translation.
\newblock {\em arXiv preprint arXiv:1609.08144 }

\bibitem[\protect\citename{Zaremba et al.}2014]{zaremba2014recurrent}
Zaremba, Wojciech, Ilya Sutskever and Oriol Vinyals.
\newblock 2014.
\newblock Recurrent neural network regularization.
\newblock {\em arXiv preprint arXiv:1409.2329}

\bibitem[\protect\citename{Zhou et al.}2016]{zhou2016minimal}
Zhou, Guo-Bing, Jianxin Wu, Chen-Lin Zhang, Zhi-Hua Zhou. 
\newblock 2016.
\newblock Minimal gated unit for recurrent neural networks.
\newblock {\em International Journal of Automation and Computing 13.3(2016)}: 226-234.

\end{thebibliography}



\end{document}